\documentclass[runningheads]{llncs}
\usepackage{amssymb}
\usepackage{pifont}%
\newcommand{\cmark}{\ding{51}}%
\newcommand{\xmark}{\ding{55}}%
\usepackage{amsfonts}
\usepackage{tabularray}
\usepackage{rotating}
\usepackage{listings}
\usepackage{multirow}
\usepackage{multicol}
\usepackage{graphics}
\usepackage{graphicx}
\usepackage{adjustbox}
\usepackage{dirtytalk}
\usepackage{algorithm}
\usepackage{algpseudocode}
\usepackage{float}
\usepackage{todonotes}
\usepackage{bm}
\usepackage{url}
\usepackage{enumerate}
\usepackage{hhline}
\usepackage{capt-of}
\usepackage[normalem]{ulem}
\usepackage{tablefootnote}
\usepackage{svg}

\begin{document}
\title{Exploring Large Language Models and Hierarchical Frameworks for Classification of Large Unstructured Legal Documents}
\titlerunning{Exploring Large Language Models and Hierarchical Frameworks}
%

\author{Nishchal Prasad\orcidID{0009-0000-4712-3540}
\and
Mohand Boughanem\orcidID{0000-0001-7004-0807}
\and
Taoufiq Dkaki\orcidID{0000-0003-3962-7663} 
}
\authorrunning{N. Prasad et al.}
%
\institute{Institut de Recherche en Informatique de Toulouse (IRIT), Toulouse, France 
\\
\email{\{Nishchal.Prasad, Mohand.Boughanem, Taoufiq.Dkaki\}@irit.fr}}

\maketitle              
\begin{abstract}
Legal judgment prediction 
suffers from the problem of long case documents exceeding tens of thousands of words, in general, and having a non-uniform structure. Predicting judgments from such documents becomes a challenging task, more so on documents with no structural annotation. We 
explore the classification of these large legal documents and their lack of structural information with a deep-learning-based hierarchical framework which we call MESc; "Multi-stage Encoder-based Supervised with-clustering”; for judgment prediction. Specifically, we divide a document into parts to extract their embeddings from the last four layers of a custom fine-tuned Large Language Model, and try to approximate their structure through unsupervised clustering. Which we use in another set of transformer encoder layers to learn the inter-chunk representations. We analyze the adaptability of Large Language Models (LLMs) with multi-billion parameters (GPT-Neo, and GPT-J) with the hierarchical framework of MESc and compare them with their standalone performance on legal texts. We also study their intra-domain(legal) transfer learning capability and the impact of combining embeddings from their last layers in MESc.
We test these methods and their effectiveness with extensive experiments and ablation studies on legal documents from India, the European Union, and the United States with the ILDC dataset and a subset of the LexGLUE dataset. Our approach achieves a minimum total performance gain of approximately 2 points over previous state-of-the-art methods.

\end{abstract}
\section{Introduction}
A legal case proceeding cycle\footnote{\url{https://www.law.cornell.edu/wex/civil_procedure}} involves analyzing vast amounts of data and legal precedents, which can be a time-consuming process given the complexity and length of the case. The number of legal cases in a country is also proportionally related to its population. This leads to a backlog of cases, especially in highly populated countries, ultimately setting back the progress of its legal system\footnote{\url{https://www.globaltimes.cn/page/202204/1260044.shtml}}\cite{katju_pending_cases}.
Automating such legal case procedures can help speed up and strengthen the decision-making process, saving time and benefiting both the legal authorities and the people involved. 
One of the fundamental problems that deal with this larger component is the prediction of the outcome based just on the case's raw texts (which can include facts, arguments, appeals, etc. except the final decision), as in a typical real-life (raw) setting. 

Several machine learning techniques have been applied to legal texts to predict judgments as a text classification problem (\cite{survey_1}, \cite{survey_2}). While it seems like a general text classification task, legal texts differ from general texts and are rather more complex, broadly in two ways, i.e. structure and syntax and, lexicon and grammar (\cite{diff_from_generalTxt_1}, \cite{legal-bert}, \cite{diff_from_generalTxt_2}). The structure of legal case documents 
is not uniform in most settings and their complex syntax and lexicon make it more difficult and expensive to annotate, 
requiring only legal professionals. This adds to another challenge of the long lengths of these documents, reaching more than 10000 words (Table \ref{dataset_des}). This lack of structure information and the long lengths of these 
legal documents pose a challenge in predicting judgments. In our work we explore this problem 
on four fronts, by (a) developing a hierarchical framework for the classification of large unstructured legal documents, (b) exploring the adaptability of billion-parameter large language models (LLMs) to this framework, (c) analyzing the performance of these LLMs without this framework and (d) checking the intra-domain(legal) transfer learning capability of domain-specific pre-trained LLMs. This is summarized below: 

\begin{itemize}
    \item We explore the problem of judgment prediction from large unstructured legal documents and propose a hierarchical multi-stage neural classification framework named \say{Multi-stage Encoder-based Supervised with-clustering} (MESc). This works by extracting embeddings from the last four layers of a fine-tuned encoder of a large language model (LLM) and using an unsupervised clustering mechanism to approximate the structure. Alongside the embeddings, these approximated structure labels are processed through another set of transformer encoder layers for final classification. 
    \item We show the effect of combining features from the last layers of transformer-based LLMs (BERT\cite{bert}, GPT-Neo\cite{gpt-neo}, GPT-J\cite{gpt-j}), along with the impact on classification upon using the approximated structure.
    \item  We study the adaptability of domain-specific pre-trained multi-billion parameter LLMs to such documents and study their intra-domain(legal) transfer learning capability (both with fine-tuning and in MESc). 
    \item 
    We performed extensive experiments and analysis on four different datasets (ILDC\cite{ildc-cjpe} and LexGLUE's \cite{lexglue} ECtHR(A), ECtHR(B), and SCOTUS) and achieved a total gain of $\approx$ 2 points in classification on these datasets.
\end{itemize}

\section{Related works}
\label{related_works}
Several strategies have been investigated 
to predict the result of legal cases in specific categories (criminal, civil, etc.) with rich annotations (Xiao et al. \cite{Chaojun}, Xu et al. \cite{xu-etal}, Zhong et al. \cite{zhong-etal-2018-legal}, Chen et al. \cite{Huajie}). These studies on well-structured and annotated legal documents show the effect and importance of having good structural information. While creating such a dataset is both time and resource (highly skilled) demanding, researchers have worked on legal documents in a more general and raw setting.
 Chalkidis et al. \cite{etchr_a} presented a dataset of European Court of Human Rights case proceedings in English, with each case assigned a score indicating its importance. They described a Legal Judgment Prediction (LJP) task, which seeks to predict the outcome of a court case using the case facts and law violations. They also create another version of this dataset \cite{ecthr_b} to give a rational explanation for the predictions. In the US legal case setting, Kaufman et al. \cite{usa_kaufman_kraft} used AdaBoost decision tree to predict the U.S. Supreme Court rulings. Tuggener et al. \cite{ledgar} proposed LEDGAR, a multilabel dataset of legal provisions in US contracts. Malik et al. \cite{ildc-cjpe} curated the Indian Legal Document Corpus (ILDC) of unannotated and unstructured documents, and used it to build baseline models for their Case Judgment Prediction and Explanation (CJPE) task 
 upon which Prasad et al. \cite{CIRCLE22} showed the possibility of intra-domain(legal) transfer learning using LEGAL-BERT on Indian legal texts.

Pretrained large language models (LLMs) based on transformers (Devlin et al.\cite{bert}, Vaswani et al.\cite{transformer}) have shown widespread success in all fields of natural language processing (NLP) but only for short texts spanning a few hundred tokens. There have been several approaches to handle longer sequences with long sequence transformer-based LLMs (Beltagy et al.\cite{longformer}, Kitaev et al.\cite{reformer}, Zaheer et al.\cite{bigbird}, Ainslie et al. \cite{ETC_transformer}). 
These architectures display similar performance as the hierarchical adaptation of their vanilla counterparts (\cite{lexglue},\cite{HAT},\cite{LSG}), and since we try to learn and approximate the structure information of the document, we choose to process the document in short sequences rather than as a whole. So, we take a different approach to handle large documents with LLMs (such as BERT \cite{bert}) based on the hierarchical idea of \say{divide, learn and combine} (Chalkidis et al.\cite{HAT}, Zhang et al. \cite{hibert}, Yang et al. \cite{hierarchical-attention-networks}), where the document is split (into parts then sentences and words, etc.) and features of each component are learned and combined hierarchically from bottom-up to get the whole document's representation. Also with the unavailability of the domain-specific pre-trained checkpoints of these long sequence LLMs and considering their expensive pretraining, we choose to use the vanilla models and develop a hierarchical adaptation.

Moreover, the domain-specific pre-training of transformer encoders has accelerated the development of NLP in legal systems with better performance as compared to the general pre-trained variants (Chalkidis et al.\cite{legal-bert}'s LEGAL-BERT trained on court cases of the US, UK, and EU, Zheng et al. \cite{case-hold}'s BERT trained on US court cases dataset CaseHOLD, Shounak et al. \cite{InLegalBERT}'s InLegalBERT and InCaseLawBERT trained on the Indian legal cases). Recently, with the emergence of multi-billion parameter LLMs such as GPT-3 \cite{GPT-3}, LLaMA \cite{llama}, LaMDA \cite{lamda}, and their superior performance in natural language understanding, researchers have tried to adapt their variants (with few-shot learning) to legal texts (Trautmann et al. \cite{gpt-legal-promt_0}, Yu et al. \cite{gpt-legal-promt_1}).
In this paper, with full-fine tuning, we check the adaptability of these billion parameter LLMs with the hierarchical framework and also their intra-domain(legal) transfer-learning compared to the intra-domain pre-training (as done in LEGAL-BERT, InLegal-BERT).
To do so we use three such variants of GPT (GPT-Neo (1.3 and 2.7)\cite{gpt-neo}, GPT-J\cite{gpt-j}) pre-trained on Pile\cite{pile}, which has a subset (FreeLaw) of court opinions of US legal cases.

\section{Method: Classification Framework (MESc)}

To handle large documents MESc architecture shares the general hierarchical idea of divide, learn, and combine (\cite{HAT}, \cite{hibert}, \cite{hierarchical-attention-networks}) but differs from the previous works in the following: 
(a) 
It uses the last four layers of the fine-tuned transformer-based LLM for extracting global representations for parts(chunks) of the document.
(b) Approximating the document structure by applying unsupervised learning (clustering) on these representations' embeddings and using this information alongside, for classification. (c) Instead of only RNNs, different combinations of transformer encoder layers are tested to get a global document representation. 
(d) Divide the process into four stages, fine-tuning, extracting embeddings, processing the embeddings (supervised + unsupervised learning), and classification.
\label{Classification Framework}

\begin{figure}
\centering
\begin{minipage}{.55\textwidth}
\centering
  \includegraphics[width=6cm]{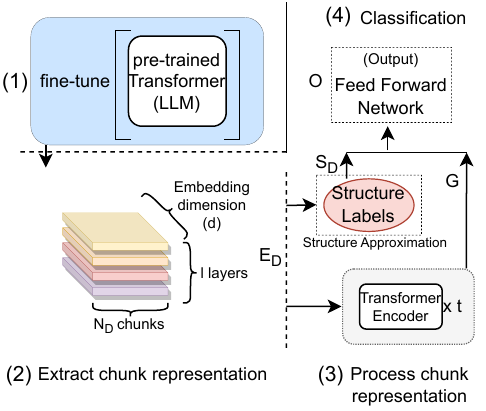}
  \caption{Multi-stage Encoder-based Supervi-\\sed with-clustering (MESc) framework.}
  \label{fig:proposed_model}
\end{minipage}%
\begin{minipage}{.45\textwidth}
\centering
    \includegraphics[height=5cm, width=4cm]{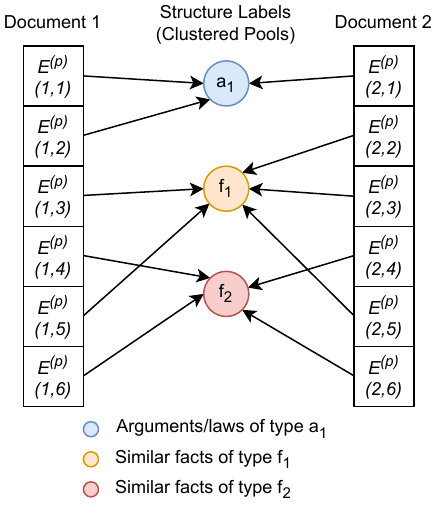}
    \caption{An example of clustering of chunk representations of two documents to generate structure labels.}
    \label{fig:cls_clustering}
\end{minipage}
\end{figure}

An overview of MESc can be seen in Fig. \ref{fig:proposed_model}. An input document $D$ is tokenized into a sequence of tokens, $D = \{t_{1,D},t_{2,D},\cdots,t_{L_D,D}\}$ via a tokenizer specific to a chosen pre-trained transformer-based language model (BERT, GPT etc.), where $t \in \mathbb{N}$ and $\mathbb{N}$ is the vocabulary of the tokenizer. This token sequence is split into a set of blocks (chunks) $\{C_{1,D},C_{2,D},\cdots,C_{N_D,D}\}$ with overlaps($o$). Where each chunk, $C_{i,D} = \{t_{((i-1)\times(c-o))},\cdots,t_{((i-1)\times(c-o)+c)}\}$ with $c$
 being the maximum number of tokens in the chunk, which is a predefined parameter for MESc (e.g. 512). $N_D = \lceil \frac{L_D}{c-o} \rceil$ is the total number of chunks for a document having $L$ tokens in total, with $o \ll c$. $N_D$ varies with the length of the document. 
 
    \paragraph{\textbf{Stage 1 - Fine-tuning:}}
    \label{stage1}
    To each chunk of a document, we associate the gold class label of the document $l_D$, and combine them to form a token matrix:
    \begin{equation} \small I_D \in \mathbb{R}^{N_D \times c \times 1} \leftarrow [\{C_{1,D},l_D\},\{C_{2,D},l_D\},\cdots,\{C_{N_D,D},l_D\}] \end{equation}
    This is used as input for the document for fine-tuning the pre-trained LLM, where $N_D$ is the batch size for one pass through the encoder. 
    This allows the encoder to adapt to the domain-specific legal texts, which helps get richer features for the next stage.
    \paragraph{\textbf{Stage 2 - Extracting chunk embeddings:}}
    \label{level(2)}
    Different layers of transformer models learn varied representations of the input sentence (\cite{bert_layer_concat_0,bert_layer_concat_1,bert_layer_concat_2}). When simultaneously used alongside each other these representations can be used to give varied features for further learning. Since the last pre-trained LLM layer captures the final representation of a chunk, we use it alongside its immediate lower layers to extract the chunk representation. We use the immediate lower layers with the intuition that the representations learned are not heavily but enough varied from the final layer.

    For a document, we pass its chunks $C_i$ through the fine-tuned encoder and extract its representation embeddings ($E_{i,D}$) from the last $l$ layers. $E_{i,D} \in \mathbb{R}^{l \times d}$, where $d$ is the dimension of the features (we use $l=4$).
    The representation embeddings can be either the first token (as in BERT) or the last token for causal language models (as in GPT). We accumulate all $E_{i,D}$ of a document to form an embedding matrix:
    \begin{equation} \small
    E_D \in \mathbb{R}^{N_D \times l \times d} \leftarrow \bigl[E_{1,D},E_{2,D},\cdots,E_{N_D,D}\bigr] 
    \end{equation}
    The $E_{i,D}$ acts as a representation of the chunk in this context, and combining them yields an approximate representation of the entire document. Doing this for all the documents gives us generated training data.

    \paragraph{\textbf{Stage 3 - Processing the extracted representations:}} \label{level(3)}
    Since the features extracted from the last layers of a fine-tuned encoder have different embedding spaces, they can contribute to give varied features. So for this stage, we choose to combine the last $p<l$ layers in $E_D$ for further training. We experiment with different $p$ before fixing one value as discussed in section \ref{results}.
    This gives $E_D^{(p)} \in \mathbb{R}^{N_D \times p \times d}, p \in \{1,2,3,4\}$.
    We used $p$=1, 2, and 4 in our experiments to compare their effects.
    We concatenate together the representations from these $p$ layers to get,
    \begin{equation} \small
    E_{i,D}^{(p)} \in \mathbb{R}^{pd \times 1} \leftarrow \bigl[E_{i,D}^{(l)}|E_{i,D}^{(l-1)}|\cdots|E_{i,D}^{(l-p-1)}\bigr] 
    \end{equation}
    This gives,
    \begin{equation} \small
    \widehat{E}_{D}^{(p)} \in \mathbb{R}^{N_D \times pd} \leftarrow \bigl\{E_{1,D}^{(p)}|E_{2,D}^{(p)}|\cdots|E_{N,D}^{(p)}\bigr\}\ 
    \end{equation}
    We also experimented with the element-wise addition of representations in $E_D^{(p)}$ and found their performance to be lower 
    in most of the experiments of section \ref{results}, hence we exclude it here.
    \begin{enumerate}[a.]
        \item \textbf{Approximating the structure labels ($S_D$)} (Unsupervised learning): \label{stage_clustering}
        To get the information on the document's structure i.e. its parts (facts, arguments, concerned laws, etc.), we use a clustering algorithm (HDBSCAN \cite{hdbscan}). We cluster the $p$ chosen extracted chunk embeddings, $\widehat{E}_{D}^{(p)}$ to map similar parts of different documents together where the labels of one part of a document are learned by its similarity with another part of another document. The idea is that the embeddings of similar parts from different documents will group forming a pool of labeled clusters that can help identify its part in the document. A synthetic example can be seen in Fig. \ref{fig:cls_clustering}, where the $E_{i,D}^{(p)}$ of documents 1 and 2 learn their cluster (label) pool for, arguments of type 
        $a_1$=\{$E^{(p)}_{1,1}, E^{(p)}_{1,2} E^{(p)}_{2,1}$\},
        facts of type 
        $f_1$=\{$E^{(p)}_{1,3},E^{(p)}_{1,5},E^{(p)}_{2,2},E^{(p)}_{2,3},E^{(p)}_{2,5}$\},
        facts of type 
        $f_2$=\{$E^{(p)}_{1,4},E^{(p)}_{1,6},E^{(p)}_{2,4},E^{(p)}_{2,6}$\}.
        So for document 1 the approximated structure then becomes $S_1$=\{$a_1$,$a_1$,$f_1$,$f_2$,$f_1$,$f_2$\} and for document 2 it is $S_2$=\{$a_1$,$f_1$,$f_1$,$f_2$,$f_1$,$f_2$\}. It is to be noticed that this distinction if it's a fact or an argument etc. is done here for representation. In an actual setting, this is unknown and the labels don't carry any specific name or meaning except for the model to give an approximation of its structure.  
        
        Since the performance of the HDBSCAN clustering mechanism decreases significantly with an increase in data dimension, we use a dimensionality reduction algorithm (pUMAP\cite{umap}), before clustering. For all the chunks of a document, their approximated structure labels are combined with the output of stage 3(b), before processing through the final classification stage (4).

        \item \textbf{Global document representation} (Supervised learning):
        For intra-chunk attention, we use transformer encoder layers (Vaswani et al. \cite{transformer}), for a chunk to attend to another through its multi-head attention and feed-forward neural network (FFN) layer. This helps the chunk representations to attend to one another in parallel. For a chunk's position in the document, we add its positional embeddings (\cite{bert}) in ${E}_{D}^{(p)}$ and process it through $t$ transformer layers $T^{(t)}_{\{h, d_f\}}$, with $h$ attention heads and $d_f = pd$ as the dimension of the FFN. $t$ and $h$ are both hyperparameters whose choice depends upon the input feature lengths. Section \ref{results} evaluates different values of these parameters, 
    but $t\geq3$ sometimes overfits the model in our experiments, hence we fix  $t=2$ for MESc. The output is max-pooled and passed through a feed-forward neural network $FFN_T$ of $128$ nodes to get:
    \begin{equation} \small
    G\left(\widehat{E}_{D}^{(p)}\right) = FFN_T\left(maxpool\left(T^{(t)}_{\{h, d_f\}}\left(\widehat{E}_{D}^{(p)}\right)\right)\right) \in \mathbb{R}^{128}
    \end{equation}
    \end{enumerate}
    \paragraph{\textbf{Stage 4 - Classification:}} \label{Level4}The structure labels along with the output of the feed-forward network of stage 3(b) are concatenated together and passed through an internal feed-forward network $FFN_i$ (32 nodes, with softmax activation) and an external feed-forward network $FFN_e$ ($u$ label/class nodes with task-specific activation function sigmoid or softmax) giving the output $O(D)$ for a document $D$ (Eq. \ref{eq.O}). $O$ and $G$ are learnt together while $S_D$ is learnt independently.
    \begin{equation} \small \label{eq.O}
    O\left(D\right) = FFN_e\left( FFN_i \left( \left(\left[G\left(\widehat{E}_{D}^{(p)}\right)|S_D\right]\right) \right) \right) \in \mathbb{R}^{u}
    \end{equation}

The code and trained models for the above implementation can be found at GitHub\footnote{\url{https://github.com/NishchalPrasad/MESc}} and our finetuned LLMs at HuggingFace\footnote{\url{https://huggingface.co/nishchalprasad}}.

\subsection{Experimental setup}
\label{Experimental setup}
Table \ref{tab:experimental_details} lists the major details for the experimental setup. For our backbone transformer-based language model, we used domain-specific models LEGAL-BERT\cite{legal-bert}, InLegalBERT\cite{InLegalBERT}, and for multi-billion parameter LLMs we chose GPT-Neo\cite{gpt-neo}, GPT-J\cite{gpt-j}. The tokenizers used are from the respective backbone transformer encoders. We abbreviate the encoders fine-tuned on 512 input length as ($\alpha$) and, for ones fine-tuned with 2048 input length as ($\gamma$). 

These hyperparameters (Table \ref{tab:experimental_details}) were used based on the guidelines of the respective language models and several of our previous experiments and dataset analyses. We list out some of them further in the paper and in discussions while referring to Table \ref{tab:finetuning_result}, Table \ref{results_table_lexglue}, Table \ref{dataset_des}, and Fig. \ref{fig:ecthr_chunk_counts}.

\begin{table}[!htb]
    \caption{Experimental setup for different stages of MESc architecture.}
    \begin{adjustbox}{width={\textwidth}}
    \label{tab:experimental_details}
    \begin{tabular}{l|l} 
    \hline
    \multicolumn{2}{l}{\begin{tabular}[c]{@{}l@{}}\uline{Stage 1}\\\textit{BERT-based LLM}: chunk-size = $512$ tokens~(90 token overlaps), [CLS] token to test.\\\textit{GPT-based LLM}: max input length=$2048$, chunk-size$\in$\{$512, 2048$\}, last token to test.\\For all ($\alpha$) GPT we compare with ($\alpha$) BERT-based LLM on $512$ input length. \\Finetuned for e=$4$ epochs, chose best e for Stage 2 and evaluation.\end{tabular}}                                                                                                                                                                                                                                                                                                                                  \\ 
    \hline
    \begin{tabular}[c]{@{}l@{}}\uline{Stage 2} (Embedding Extraction)\textit{}\\\textit{}\\\textit{BERT-based LLM:~}\\{[}CLS] token for each chunk\\\textit{}\\\textit{GPT-based LLM:}\\last token for each chunk\end{tabular} & \begin{tabular}[c]{@{}l@{}}\uline{Stage 3 \& 4}\\\textit{Optimizer} = Adam (learning rate = $3.5e^{-6}$)\\\textit{Loss func.:~}multi-label: categorical cross-entropy\\~ ~ ~ ~ ~ ~ ~ ~ binary \& multi-class: binary cross-entropy\\$t$=\{$1,2,3$\}, $h$=$8$, e=$5$ epochs (best e for evaluation) \\\textit{Structure approximation:} pUMAP\tablefootnote{\url{https://umap-learn.readthedocs.io/en/latest/parametric_umap.html}} (64 dimensions)\\~ ~ ~ ~ ~ ~ ~ ~ ~ ~ ~ ~ ~ ~ ~ ~ ~HDBSCAN ($15$ min clusters)\end{tabular}  \\ 
    \hline
    \multicolumn{2}{l}{\begin{tabular}[c]{@{}l@{}}GPU used: Nvidia V100 \& A100, with ZERO-3 in Deepspeed\tablefootnote{\url{https://www.deepspeed.ai/}} with Accelerate\tablefootnote{\url{https://huggingface.co/docs/accelerate/index}}.\\Maximum fine-tune time (hours/epoch) for GPTs (6 Nvidia A100): \\GPT-Neo-1.3B = 2.1, GPT-Neo-2.7B = 4, GPT-J = 8\end{tabular}}                                                                                                                                                                                                                                                                                                                                                                                                       \\
    \hline
    \end{tabular}
    \end{adjustbox}
    \\
    \\
    \begin{tabular}{cc}
        \begin{minipage}{0.5\textwidth} 
           \centering
            \caption{Dataset statistics}
            \begin{adjustbox}{width=\columnwidth}
            \label{dataset_des}
            \begin{tabular}{c|c|c|c|c|c}
            \multicolumn{2}{c|}{Name}                                                                         & \textbf{ILDC}                                        & \textbf{ECtHR(A)}                                     & \textbf{ECtHR(B)}                                     & \textbf{SCOTUS}                                        \\ 
            \hline
            \multirow{3}{*}{\begin{tabular}[c]{@{}c@{}}No. of~\\Docs.\end{tabular}}                   & Train & 37387                                                & 9000                                                  & 9000                                                  & 5000                                                   \\ 
            \cline{2-6}
                                                                                                      & Val.  & 994                                                  & 1000                                                  & 1000                                                  & 1400                                                   \\ 
            \cline{2-6}
                                                                                                      & Test  & 1517                                                 & 1000                                                  & 1000                                                  & 1400                                                   \\ 
            \hline
            \multirow{3}{*}{\begin{tabular}[c]{@{}c@{}}Average~\\tokens\\~\\Max\\tokens\end{tabular}} & Train & \begin{tabular}[c]{@{}c@{}}4120\\501275\end{tabular} & \begin{tabular}[c]{@{}c@{}}2011\\46500\end{tabular}   & \begin{tabular}[c]{@{}c@{}}2011\\46500\end{tabular}   & \begin{tabular}[c]{@{}c@{}}8291\\126377\end{tabular}   \\ 
            \cline{2-6}
                                                                                                      & Val.  & \begin{tabular}[c]{@{}c@{}}8048\\51045\end{tabular}  & \begin{tabular}[c]{@{}c@{}}2210\\18352\end{tabular}   & \begin{tabular}[c]{@{}c@{}}2210\\18352\end{tabular}   & \begin{tabular}[c]{@{}c@{}}12639\\56310\end{tabular}   \\ 
            \cline{2-6}
                                                                                                      & Test  & \begin{tabular}[c]{@{}c@{}}5238\\55703\end{tabular}  & \begin{tabular}[c]{@{}c@{}}2401\\20835\end{tabular}   & \begin{tabular}[c]{@{}c@{}}2401\\20835\end{tabular}   & \begin{tabular}[c]{@{}c@{}}12597\\124955\end{tabular}  \\ 
            \hline
            \multicolumn{2}{c|}{No. of labels}                                                                & 2                                                    & 10                                                    & 10                                                    & 13                                                     \\ 
            \hline
            \multicolumn{2}{c|}{Problem Type}                                                                 & Binary                                               & \begin{tabular}[c]{@{}c@{}}Multi-\\Label\end{tabular} & \begin{tabular}[c]{@{}c@{}}Multi-\\Label\end{tabular} & \begin{tabular}[c]{@{}c@{}}Multi-\\Class\end{tabular}  \\
            \hline
            \end{tabular}
            \end{adjustbox}
        \end{minipage}& 
        \begin{minipage}{0.5\textwidth} 
            \centering
            \includegraphics[height=4cm, width=0.9\columnwidth]{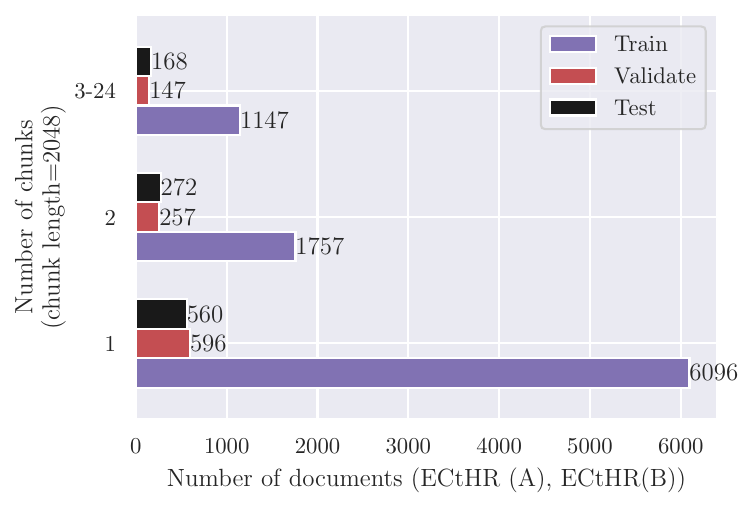} 
            \captionof{figure}{Number of documents vs. the number of chunks for ECtHR.}
            \label{fig:ecthr_chunk_counts}
        \end{minipage}
    \end{tabular}
    
    \caption{Class distribution of the datasets.}
    \begin{adjustbox}{width={\textwidth}}
    \label{tab:class_distribution}
    \begin{tabular}{|c|c|c|c|c|c|c|c|c|c|c|c|c|c|c|}
    \multicolumn{1}{c|}{(Problem type)}                                                         & \multicolumn{14}{c}{class : \# documents}                                                                                                        \\ 
    \hline
    \multirow{3}{*}{\begin{tabular}[c]{@{}c@{}}\textbf{ILDC }\\(Binary)\end{tabular}}           & Train & 0: 22067 & 1: 15320 & \multicolumn{11}{c}{\multirow{3}{*}{}}                                                                             \\ 
    \cline{2-4}
                                                                                                & Val.  & 0: 497   & 1: 497   & \multicolumn{11}{c}{}                                                                                              \\ 
    \cline{2-4}
                                                                                                & Test  & 0: 755   & 1: 762   & \multicolumn{11}{c}{}                                                                                              \\ 
    \cline{1-12}
    \multirow{3}{*}{\begin{tabular}[c]{@{}c@{}}\textbf{ECtHR (A) }\\(Multi-label)\end{tabular}} & Train & 0: 505   & 1: 1349  & 2: 1368 & 3: 4704 & 4: 710  & 5: 41 & 6: 291 & 7: 110  & 8: 141 & 9: 1421 & \multicolumn{3}{c}{\multirow{6}{*}{}}  \\ 
    \cline{2-12}
                                                                                                & Val.  & 0: 57    & 1: 193   & 2: 187  & 3: 300  & 4: 87   & 5: 4  & 6: 42  & 7: 33   & 8: 18  & 9: 139  & \multicolumn{3}{c}{}                   \\ 
    \cline{2-12}
                                                                                                & Test  & 0: 56    & 1: 189   & 2: 166  & 3: 299  & 4: 123  & 5: 5  & 6: 77  & 7: 37   & 8: 16  & 9: 122  & \multicolumn{3}{c}{}                   \\ 
    \cline{1-12}
    \multirow{3}{*}{\begin{tabular}[c]{@{}c@{}}\textbf{ECtHR (B) }\\(Multi-label)\end{tabular}} & Train & 0: 623   & 1: 1740  & 2: 1623 & 3: 5437 & 4: 1056 & 5: 81 & 6: 441 & 7: 162  & 8: 444 & 9: 1558 & \multicolumn{3}{c}{}                   \\ 
    \cline{2-12}
                                                                                                & Val.  & 0: 75    & 1: 236   & 2: 219  & 3: 394  & 4: 153  & 5: 9  & 6: 64  & 7: 39   & 8: 34  & 9: 168  & \multicolumn{3}{c}{}                   \\ 
    \cline{2-12}
                                                                                                & Test  & 0: 76    & 1: 234   & 2: 196  & 3: 394  & 4: 188  & 5: 11 & 6: 106 & 7: 43   & 8: 32  & 9: 155  & \multicolumn{3}{c}{}                   \\ 
    \hline
    \multirow{3}{*}{\begin{tabular}[c]{@{}c@{}}\textbf{SCOTUS }\\(Multi-class)\end{tabular}}    & Train & 0: 1011  & 1: 811   & 2: 423  & 3: 193  & 4: 45   & 5: 35 & 6: 255 & 7: 1043 & 8: 717 & 9: 191  & 10: 53 & 11: 220 & 12: 3               \\ 
    \cline{2-15}
                                                                                                & Val.  & 0: 360   & 1: 218   & 2: 108  & 3: 70   & 4: 22   & 5: 35 & 6: 51  & 7: 226  & 8: 165 & 9: 83   & 10: 14 & 11: 38  & 12: 10              \\ 
    \cline{2-15}
                                                                                                & Test  & 0: 372   & 1: 222   & 2: 88   & 3: 51   & 4: 28   & 5: 17 & 6: 24  & 7: 260  & 8: 200 & 9: 83   & 10: 15 & 11: 37  & 12: 3               \\
    \hline
    \end{tabular}
    \end{adjustbox}
\end{table}

\subsection{Dataset:}
\label{dataset_desc}

We chose the legal datasets having large documents with a nonuniform structure throughout and without any structural annotations. The ILDC dataset \cite{ildc-cjpe} includes highly unstructured 39898 English-language case transcripts from the Supreme Court of India (SCI), where the final decisions have been removed from the document. Upon analyzing the documents from their sources and the dataset we found that they are highly unstructured and noisy. The initial decision between "rejected" and "accepted" made by the SCI judge(s) is used to classify each document and serves as their decision label. The LexGLUE dataset \cite{lexglue} comprises a set of seven datasets from the European Union and US court case setting, for uniformly assessing model performance across a range of legal NLP tasks, from which we choose ECtHR (Task A), ECtHR (Task B), and SCOTUS as they are classification tasks involving long unstructured legal documents. ECtHR (A and B) are court cases from the European Convention on Human Rights (ECHR) for articles that were violated or allegedly violated. The dataset contains factual paragraphs from the description of the cases. SCOTUS consists of court cases from the highest federal court in the United States of America, with metadata from SCDB \footnote{\url{http://scdb.wustl.edu/}}. The details of the number of labels, the document lengths (in tokens), task description, and class distribution can be found in Table \ref{dataset_des} and Table \ref{tab:class_distribution}. The tokenization Table \ref{dataset_des} is done using the tokenizer of GPT-J. 

For performance comparison on LexGLUE, we used the SOTA benchmark of Chalkidis et al. \cite{lexglue}, Condevaux et al.'s LSG \cite{LSG}, Chalkidis et al.'s HAT \cite{HAT} and for ILDC we used its benchmark from \cite{ildc-cjpe} and of Shounak et al. \cite{InLegalBERT}'s experiments.

\section{Results and discussion}
\label{results}
$\mu$-F$1$ (micro) and $m$-F$1$ (macro) are used to measure the performance for the LexGLUE dataset, and accuracy(\%) and macro-F$1$ for the ILDC dataset. These metrics were chosen partly to compare with the previous benchmark models (stated in Table \ref{Performance with MESc}) conforming to their original results and metrics. 
We list out the detailed experimental results for best configurations of MESc in Table \ref{results_table_lexglue} 
and the fine-tuned performance of the LLMs used in Table \ref{tab:finetuning_result}. 
\paragraph{\textbf{Intra-domain(legal) transfer learning:}} 
Based on the analysis of ILDC by Malik et al. \cite{ildc-cjpe} we use the last chunk of the documents for evaluation. As can be seen from Table \ref{tab:finetuning_result}, for LexGLUE's subset, all the GPTs used here adapt better than the BERT-based models with a minimum of $\approx$ 3 points gain on $\mu$-F1 and a minimum of $\approx$ 6 points on m-F1 score.
On the other hand in the ILDC dataset, for the $\alpha$ variants with 512 input lengths for evaluation, the performance dropped or remained similar to the InLegalBERT, while upon increasing the evaluation input length to 2048 we can see an increase of more than 1 point in the performance. When fine-tuned with 2048 input length, the performance of GPT-J ($\gamma$) compared to its $\alpha$ and $\beta$ variant is at least $\approx$ 2 points higher for all the datasets. We can see that an increase in the input length for fine-tuning helps to capture more feature information for such documents. Also going from GPT-Neo-1.3B to GPT-Neo-2.7B to GPT-J-6B, the performance increases by a margin of 2 points at minimum, here we see the parameter count playing an important role in adapting and understanding these documents. Even though GPT-Neo and GPT-J are pre-trained on US legal cases (Pile \cite{pile}) they adapt better to the European and Indian legal documents, with a minimum gain of $\approx$ 7 points ($\gamma$) on the ECtHR(A \& B) and the ILDC dataset over their domain-specific pre-trained counterparts LEGAL-BERT and InLegalBERT respectively. These results show the transfer learning capacity of LLMs between different legal domains with different settings, which can be a better alternative with limited resources compared to expensive domain-specific pre-training. 

\begin{table}
    \caption{Fine-tuned results on the last chunk for the chosen LLMs (Section \ref{Experimental setup})}
    \begin{adjustbox}{width={\textwidth}}\small
    \label{tab:finetuning_result}
    \begin{tabular}{c|c|c|c|c|c} 
    \hline
    \multicolumn{6}{l}{\begin{tabular}[c]{@{}l@{}}$\alpha$: fine-tuned and evaluated with 512 input length, $\beta$: evaluating $\alpha$ on its maximum input length,\\$\gamma$: fine-tuned and evaluated with its maximum input length. All measures are in (\%). e = epoch.\end{tabular}}                                                                                                                                                                                                                                                                                                                \\ 
    \hline
    \multicolumn{2}{c|}{\textbf{Dataset}}                                                                                                                & \begin{tabular}[c]{@{}c@{}}\textbf{LEGAL-BERT}\\($\mu$-F1/m-F1)\end{tabular}               & \begin{tabular}[c]{@{}c@{}}\textbf{GPT-Neo 1.3B}\\($\mu$-F1/m-F1)\end{tabular}                                            & \begin{tabular}[c]{@{}c@{}}\textbf{GPT-Neo 2.7B}\\($\mu$-F1/m-F1)\end{tabular}                                            & \begin{tabular}[c]{@{}c@{}}\textbf{GPT-J 6B}\\($\mu$-F1/m-F1)\end{tabular}                                                          \\ 
    \hline
    \multirow{3}{*}{\begin{tabular}[c]{@{}c@{}}\\~\\~\\~\\LexGLUE's\\subset\end{tabular}} & \begin{tabular}[c]{@{}c@{}}ECtHR \\(A)\end{tabular} & \begin{tabular}[c]{@{}c@{}}($\alpha$)\\62.85/48.66\\(e = 4)\end{tabular}          & \begin{tabular}[c]{@{}c@{}}($\alpha$)66.19/56.59\\($\beta$)66.20/57.16\\(e = 2)\end{tabular}                       & \begin{tabular}[c]{@{}c@{}}($\alpha$)68.49/54.45\\($\beta$)68.11/56.49\\(e = 2)\end{tabular}                       & \begin{tabular}[c]{@{}c@{}}($\alpha$)71.42/59.27\\($\beta$)73.30/62.45\\($\gamma$)\textbf{74.51/64.67}\\(e = 3)\end{tabular}  \\ 
    \cline{2-6}
                                                                                          & \begin{tabular}[c]{@{}c@{}}ECtHR\\(B)\end{tabular}  & \begin{tabular}[c]{@{}c@{}}($\alpha$)\\70.89/64.05\\(e = 3)\end{tabular}          & \begin{tabular}[c]{@{}c@{}}($\alpha$)75.42/70.91\\($\beta$)75.74/70.09\\(e = 2)\end{tabular}                       & \begin{tabular}[c]{@{}c@{}}($\alpha$)74.48/68.26\\($\beta$)75.13/70.72\\(e = 2)\end{tabular}                       & \begin{tabular}[c]{@{}c@{}}($\alpha$)77.15/73.26\\($\beta$)80.49/76.31\\($\gamma$)\textbf{83.16/79.27}\\(e = 3)\end{tabular}  \\ 
    \cline{2-6}
                                                                                          & SCOTUS                                              & \begin{tabular}[c]{@{}c@{}}($\alpha$)\\68.76/53.57\\(e = 6)\end{tabular}          & \begin{tabular}[c]{@{}c@{}}($\alpha$)71.14/60.35\\($\beta$)73.71/63.10\\($\gamma$)75.02/64.38\\(e = 2)\end{tabular} & \begin{tabular}[c]{@{}c@{}}($\alpha$)70.57/60.25\\($\beta$)73.64/65.64\\($\gamma$)76.36/66.19\\(e = 1)\end{tabular} & \begin{tabular}[c]{@{}c@{}}($\alpha$)72.00/62.76\\($\beta$)75.71/66.25\\($\gamma$)\textbf{78.50/71.96}\\(e = 3)\end{tabular}  \\ 
    \hline
    \multicolumn{1}{l}{}                                                                  & \multicolumn{1}{l|}{}                               & \textbf{InLegalBERT} (Acc./m-F1)                                                           & \multicolumn{3}{c}{Accuracy (Acc.) / m-F1}                                                                                                                                                                                                                                                                                                                       \\ 
    \cline{3-6}
    \multicolumn{2}{c|}{ILDC}                                                                                                                   & \begin{tabular}[c]{@{}c@{}}($\alpha$)\\\textbf{76.00/76.10}\\(e = 4)\end{tabular} & \begin{tabular}[c]{@{}c@{}}($\alpha$)72.91/72.91\\($\beta$)77.26/77.25\\(e=1)\end{tabular}                         & \begin{tabular}[c]{@{}c@{}}($\alpha$)74.29/74.24\\($\beta$)81.21/81.18\\(e=1)\end{tabular}                         & \begin{tabular}[c]{@{}c@{}}($\alpha$)73.96/73.96\\($\beta$)81.93/81.92\\($\gamma$)\textbf{83.72/83.66}\\(e=1)\end{tabular}    \\
    \hline
    \end{tabular}
    \end{adjustbox}
    \\
    \\
    \caption{Test results for different configurations of MESc. We show the maximum scores attained in all the runs. The bold-faced values also signify statistically significant findings in 5 different runs. (The baseline results are from their original papers.)}
    \centering
    \begin{adjustbox}{width={\columnwidth}}
    \label{results_table_lexglue}
    \begin{tabular}{|c|l|c|c|c|c|c|c|} 
    \hline
    \multicolumn{8}{c}{\begin{tabular}[c]{@{}c@{}}* is the fine-tuned LLM used for embedding extraction (Table \ref{tab:finetuning_result}).\\$p$ = last $p$ layers of the * model; $t$ transformer encoder layers; $S_D$ = approximated structure.\end{tabular}}                                                                                                                                \\ 
    \hline
    \multicolumn{1}{l}{}                                                                & \multicolumn{2}{l|}{}                                  & \textbf{ECtHR (A)}   & \textbf{ECtHR (B)}   & \textbf{SCOTUS}      & \multicolumn{1}{l|}{}                                                               & \textbf{ILDC}                                 \\ 
    \cline{4-6}\cline{8-8}
    \multicolumn{1}{l}{}                                                                & \multicolumn{2}{c|}{}                                  & \multicolumn{3}{c|}{\textbf{(\%) $\mu$-F1/m-F1}}                   & \multicolumn{1}{l|}{}                                                               & \multirow{2}{*}{\textbf{(\%) Accuracy/m-F1}}  \\ 
    \cline{1-6}
    \multicolumn{3}{c|}{Chalkidis et al.\cite{lexglue}}                                                                         & 71.2/64.7            & 80.4/74.7            & 76.6/66.5            & \multicolumn{1}{l|}{}                                                               &                                               \\ 
    \hline
    \multicolumn{3}{c|}{LSG \cite{LSG}}                                                                                         & 71.7/63.9            & 81.0/75.1            & 74.5/62.6            & Malik et al.\cite{ildc-cjpe}                                       & 77.78/77.79                                   \\ 
    \hline
    \multicolumn{3}{c|}{HAT \cite{HAT}}                                                                                         & -                    & 80.8/79.8            & -                    & Shounak et al.\cite{InLegalBERT}                                   & -/83.09                                       \\ 
    \hline
    \multicolumn{3}{c|}{\textbf{MESc}}                                                                                                           & \multicolumn{1}{c}{} & \multicolumn{1}{c}{} &                      & \multicolumn{1}{l|}{}                                                               &                                               \\ 
    \cline{1-3}
                                                                                        & \multicolumn{1}{c|}{$p$,~$t$}  & $S_D$                 & \multicolumn{1}{c}{} & \multicolumn{1}{c}{} &                      &                                                                                     &                                               \\ 
    \hline
    \multirow{10}{*}{\begin{tabular}[c]{@{}c@{}}LEGAL-BERT* \\($\alpha$)\end{tabular}}  & \multirow{2}{*}{$p$=1, $t$=1~} & \xmark & 68.25/58.06          & 74.18/68.90          & 71.36/59.16          & \multirow{10}{*}{\begin{tabular}[c]{@{}c@{}}InLegalBERT* \\($\alpha$)\end{tabular}} & 83.72/83.73                                   \\
                                                                                        &                                & \cmark & -                    & -                    & -                    &                                                                                     & 83.65/83.65                                   \\ 
    \cline{2-6}\cline{8-8}
                                                                                        & \multirow{2}{*}{$p$=1, $t$=2}  & \xmark & 69.23/59.35          & 73.86/67.42          & 71.52/58.17          &                                                                                     & 83.45/83.47                                   \\
                                                                                        &                                & \cmark & -                    & -                    & -                    &                                                                                     & 83.78/83.78                                   \\ 
    \cline{2-6}\cline{8-8}
                                                                                        & \multirow{2}{*}{$p$=4, $t$=1}  & \xmark & 75.46/62.26          & 81.02/75.73          & 73.96/58.65          &                                                                                     & 83.41/83.41                                   \\
                                                                                        &                                & \cmark & 75.82/63.78          & 81.22/77.25          & 75.25/61.94          &                                                                                     & \textbf{\textbf{84.15/84.15}}                 \\ 
    \cline{2-6}\cline{8-8}
                                                                                        & \multirow{2}{*}{$p$=4, $t$=2}  & \xmark & 75.43/63.37          & 81.18/75.64          & 74.31/60.54          &                                                                                     & 83.72/83.68                                   \\
                                                                                        &                                & \cmark & \textbf{76.18/65.08} & \textbf{81.57/76.70} & \textbf{75.50/62.08} &                                                                                     & \textbf{\textbf{84.11/84.13}}                 \\ 
    \cline{2-6}\cline{8-8}
                                                                                        & \multirow{2}{*}{$p$=4, $t$=3}  & \xmark & 75.23/63.11          & 81.32/76.99          & 73.99/56.35          &                                                                                     & -                                             \\
                                                                                        &                                & \cmark & 75.10/63.09          & 81.00/76.21          & 73.92/57.83          &                                                                                     & -                                             \\ 
    \hline
    \multirow{4}{*}{\begin{tabular}[c]{@{}c@{}}GPT-Neo 1.3B* \\($\alpha$)\end{tabular}} & \multirow{2}{*}{$p$=2, $t$=2}  & \xmark & 71.15/63.59          & 80.30/77.02          & 75.36/64.79          & \multicolumn{1}{l|}{\multirow{4}{*}{}}                                              & -                                             \\
                                                                                        &                                & \cmark & \textbf{72.73/64.48} & \textbf{80.40/78.08} & \textbf{76.46/65.92} & \multicolumn{1}{l|}{}                                                               & -                                             \\ 
    \cline{2-6}\cline{8-8}
                                                                                        & \multirow{2}{*}{$p$=4, $t$=2}  & \xmark & 71.46/62.77          & 80.86/76.64          & 74.29/63.52          & \multicolumn{1}{l|}{}                                                               & -                                             \\
                                                                                        &                                & \cmark & 70.68/64.10          & 80.60/77.57          & 74.18/63.77          & \multicolumn{1}{l|}{}                                                               & -                                             \\ 
    \hline
    \multirow{4}{*}{\begin{tabular}[c]{@{}c@{}}GPT-Neo 2.7B* \\($\alpha$)\end{tabular}} & \multirow{2}{*}{$p$=2, $t$=2}  & \xmark & 74.57/62.24          & 79.49/76.20          & 76.76/65.70          & \multirow{4}{*}{\begin{tabular}[c]{@{}c@{}}GPT-Neo 2.7B*\\($\alpha$)\end{tabular}}  & 82.97/82.79                                   \\
                                                                                        &                                & \cmark & \textbf{75.67/66.44} & \textbf{80.72/76.96} & \textbf{76.27/66.30} &                                                                                     & 83.65/83.64                                   \\ 
    \cline{2-6}\cline{8-8}
                                                                                        & \multirow{2}{*}{$p$=4, $t$=2}  & \xmark & 75.24/63.55          & 79.40/75.03          & 75.77/65.54          &                                                                                     & 83.01/83.00                                   \\
                                                                                        &                                & \cmark & 75.87/65.61          & 79.35/76.35          & 76.41/67.75          &                                                                                     & 83.22/83.21                                   \\ 
    \hline
    \multirow{4}{*}{\begin{tabular}[c]{@{}c@{}}GPT-J 6B* \\($\alpha$)\end{tabular}}     & \multirow{2}{*}{$p$=2, $t$=2}  & \xmark & 72.22/62.63          & 79.31/76.92          & 75.05/66.58          & \multirow{4}{*}{\begin{tabular}[c]{@{}c@{}}GPT-J 6B*\\($\alpha$)\end{tabular}}      & 82.84/82.78                                   \\
                                                                                        &                                & \cmark & \textbf{71.63/64.06} & \textbf{79.77/77.60} & \textbf{75.98/67.15} &                                                                                     & 83.21/83.19                                   \\ 
    \cline{2-6}\cline{8-8}
                                                                                        & \multirow{2}{*}{$p$=4, $t$=2}  & \xmark & 71.56/61.18          & 78.00/76.05          & 74.90/63.33          &                                                                                     & 82.73/82.73                                   \\
                                                                                        &                                & \cmark & 72.19/64.37          & 77.95/76.25          & 74.85/65.93          &                                                                                     & 83.37/83.36                                   \\ 
    \hline
    \multirow{4}{*}{\begin{tabular}[c]{@{}c@{}}GPT-J 6B*\\($\gamma$)\end{tabular}}      & \multirow{2}{*}{$p$=2, $t$=2}  & \xmark & 73.84/64.34          & 80.94/76.75          & 76.88/67.73          & \multicolumn{1}{l}{}                                                                & \multicolumn{1}{c}{}                          \\
                                                                                        &                                & \cmark & \textbf{74.70/65.71} & \textbf{81.69/78.01} & 78.14/68.53          & \multicolumn{1}{l}{}                                                                & \multicolumn{1}{c}{}                          \\ 
    \cline{2-6}
                                                                                        & \multirow{2}{*}{$p$=4, $t$=2}  & \xmark & 72.96/63.33          & 81.13/77.63          & 77.28/67.86          & \multicolumn{1}{l}{}                                                                & \multicolumn{1}{c}{}                          \\
                                                                                        &                                & \cmark & 74.84/65.48          & 81.34/78.02          & \textbf{78.67/69.66} & \multicolumn{1}{l}{}                                                                & \multicolumn{1}{c}{}                          \\
    \cline{1-6}
    \end{tabular}
    \end{adjustbox}
\end{table}

\paragraph{\textbf{Performance with MESc:}}
\label{Performance with MESc}
Looking at Table \ref{results_table_lexglue} we interpret the results in two directions.

    \textbf{(a) Encoders fine-tuned on 512 input length $(\alpha)$:} 
    For LEGAL-BERT and InLegalBERT in all datasets, MESc achieves a significant increase in performance by at least 
    4 points in all metrics than their fine-tuned LLM counterparts with just the last layer ($p$=1). Combining the last four layers with $t$=1 encoder layer 
    yields a performance boost of 4 points or more in ECtHR datasets while there is not much improvement in ILDC and SCOTUS. With $S_D$, the approximated structure labels, there is a slight performance increase in the ILDC. 
    The same goes for SCOTUS with $\approx$ 1 point increase. 
    With the same configuration and $t$=2 encoder layers, 
    we can see a much bigger performance with the structure labels achieving new baseline scores in ECtHR (A), ECtHR (B), and ILDC datasets. For SCOTUS, this improvement from the baseline is not much. 
    This is because of the high skew of class labels in the test dataset (for example label 5 has only 5 samples). With these results, we fixed 
    certain parameters in MESc for further experiments with the extracted embeddings from GPT-Neo and GPT-J. For them, we ran experiments with $t$=2 encoders 
    and the last layer ($p$=1) 
    and gained lesser performance than $p$=2 (or 4) layers and $t$=2 
    encoders, which we exclude in this paper. For ECtHR(A\&B) and SCOTUS, concatenating the embeddings from the last two layers of GPT-Neo or GPT-J had a significant impact above their vanilla fine-tuned variants by a minimum margin of 3 points for GPT-Neo-1.3B, and 1 point for GPT-Neo-2.7B and GPT-J. This increases further by a minimum of 1 point when including the approximated structure labels, showing the impact of having structural information. 
    For ILDC, concatenating the last four layers didn't have much improvement in the performance while including the generated structure labels increased the performance. 
    
    \textbf{(b) Encoders fine-tuned on 2048 input length $(\gamma)$:} 
    Referring to Table \ref{tab:finetuning_result} and Table \ref{results_table_lexglue} for the documents we did a comparative study of MESc(on GPT-J 6B* ($\gamma$))'s performance with its backbone fine-tuned LLM (GPT-J 6B ($\gamma$)) to see the effect of increasing the number of parameters and the input length. GPT-J 6B ($\gamma$) fine-tuned on its maximum input length (2048) achieves better (or similar) performance than its MESc overhead trained on its extracted embeddings. For SCOTUS, MESc achieves better performance (2 points, m-F1) in the test set. Almost similar performance (m-F1) in ECtHR(B), 1 point higher (m-F1) in ECtHR(A)'s test set, and lesser in ILDC. 
    To check if this is the case with GPT-Neo-1.3B and GPT-Neo-2.7B we fine-tuned them with their maximum input length (2048) on SCOTUS (which through our experiments can be seen are more difficult to classify). We found that fine-tuning GPT-Neo (1.3B and 2.7B) on its maximum input length didn't show the same results as with the GPT-J. We find that for both GPT-Neo-1.3B($\gamma$) and GPT-Neo-2.7B($\gamma$) even the MESc (GPT-Neo-1.3B*($\alpha$)) and MESc (GPT-Neo-2.7B*($\alpha$)) performs better ($>$ 1 point m-F1) respectively. 
    To analyze this, we plot the distribution of the number of documents with respect to their chunk counts (chunk length = 2048) in the datasets, one such example of ECtHR can be found in Fig. \ref{fig:ecthr_chunk_counts} (we accumulate the document counts for chunks 3 to 24 for clarity). 
    As observed, most of the documents can fit in 1 or 2 chunks (median = 1), which means that with the longer input of 2048, most of the important information is not fragmented during the fine-tuning process (stage 1) and prediction. Along with this, the higher number of parameters in GPT-J helps it adapt better to most of the documents. 
    We observe that most ($>$ 90\%) of the documents can fit in very few chunks, deepening the models with extra layers (stages 3 \& 4) does not have any added value.

    The results obtained are statistically significant\footnote{We performed student's t-test (p-value $< 0.05$).} \cite{Welch_statistical_significance}.
\begin{figure}
  \centering
  \includegraphics[width=\columnwidth, keepaspectratio]{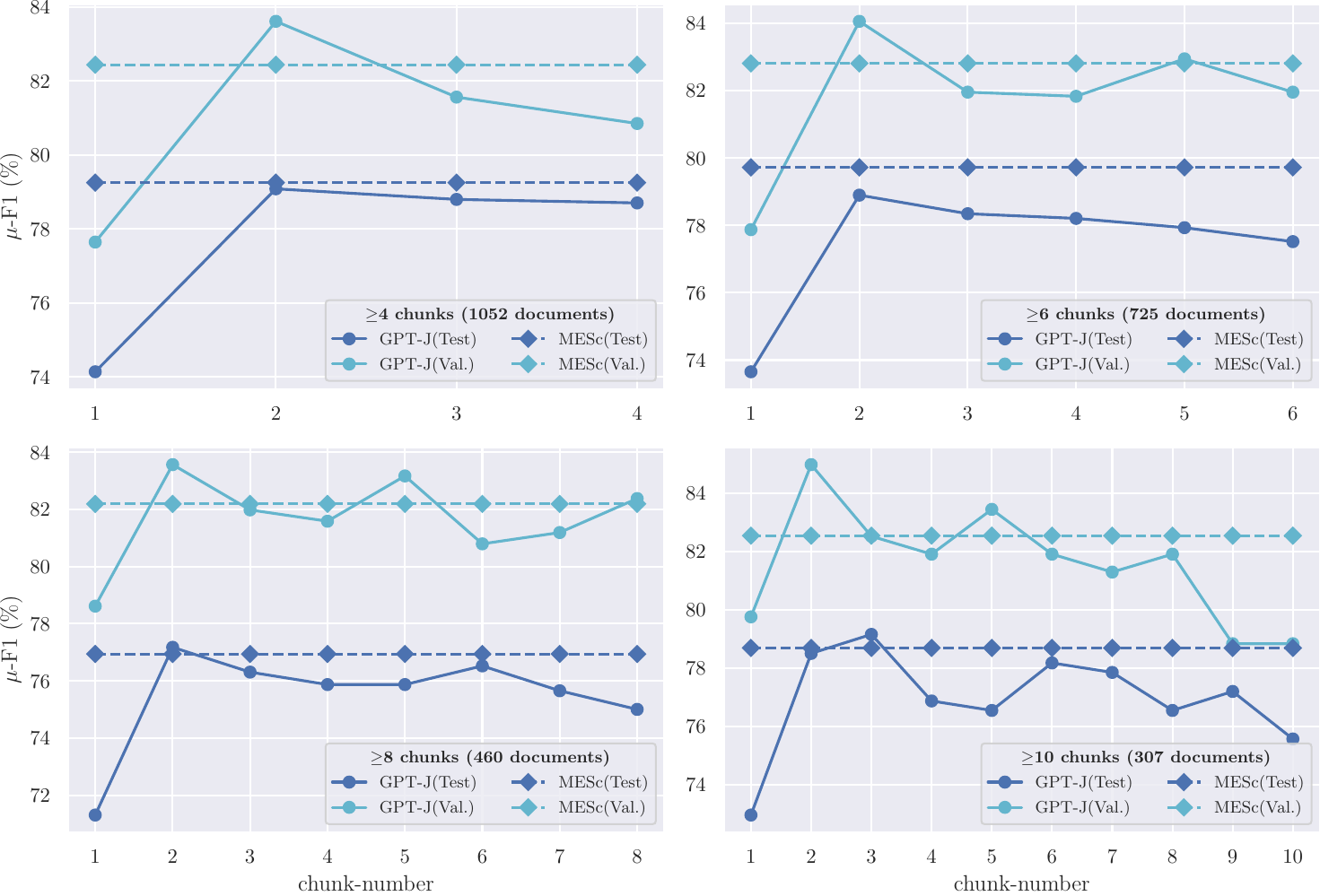} 
  \caption{$\mu$-F1 for chunk-number for GPT-J ($\gamma$) vs MESc (GPT-J* ($\gamma$)) in SCOTUS on both Validation(Val.) and Test set.}
  \label{fig:chunk_number_vs_prediction_score}
\end{figure}

\paragraph{\textbf{Analysis on long documents:}} To analyze the performance of MESc and its standalone LLM with the document length we ran experiments with GPT-J ($\gamma$) on documents with minimum lengths of 4, 6, 8, and 10 chunks of the documents. The predictions are made using the $n$th chunk of all the documents. We show these results from the SCOTUS dataset in Fig. \ref{fig:chunk_number_vs_prediction_score}, where, for the input chunk of the documents, we plot its corresponding $\mu$-F1 score. Since MESc has no input length limit and takes all the chunks at once we plot its prediction for all chunks considered as constant lines. The performance with GPT-J ($\gamma$) fluctuates with the input chunk, with the worst performance when using the first chunk and the best on the second/third chunk. This shows that for these documents in the test set the second/third chunk has a higher probability of containing the important information for a more robust prediction. Choosing which chunk to use for an unseen test set becomes more difficult as the document length increases and there is no prior information on its important parts. The fluctuations become worse for documents with a minimum of 10 chunks. While for MESc the performance is overall better than GPT-J ($\gamma$) in all the lengths considered. This shows that the hierarchical framework (such as MESc) is more reliable than its LLM counterpart on longer documents and when the important parts of the document are unknown.

With these results on MESc, we find that: 

    1. Concatenating embeddings from the last two layers in GPT-Neo (1.3B, 2.7B) or GPT-J or, the last 4 layers in BERT-based models, provides the optimum number of feature variances. 
    Globally, concatenating the embeddings helped to get a better approximation of the structure labels and improved performance.
    
    2. MESc works better than its counterpart LLM under the condition that the length of most of the documents in the dataset is much greater than the maximum input length of the LLM.

    3. For long documents when their important parts are unknown MESc performs better than its counterpart LLMs.

\section{Conclusion}
We 
explore the problem of classification of large and unstructured legal documents and develop a multi-stage hierarchical classification framework (MESc). 
We test the effect of including our approximated structure and the impact of combining the embeddings from the last layers of a fine-tuned transformer-based LLM in MESc. Along with BERT-based LLMs, we explored the adaptability of LLMs with billion parameters (GPT-Neo and GPT-J) to MESc and analyzed its limits (section \ref{Performance with MESc}) with these LLMs suggesting 
the optimal condition for its performance. The benchmark performance of GPT-Neo and GPT-J on the legal cases from India and Europe shows the intra-domain(legal) transfer learning capability of these billion-parameter language models. Most of all, our experiments achieve a new baseline in the classification of the ILDC and the LexGLUE subset (ECtHR (A), ECtHR (B), and SCOTUS). In our future work, we aim to analyze the clusters and how they contribute to the prediction. We aim to develop an explanation algorithm to explain the predictions while also leveraging this work in-domain, on the French and European legal cases to further explore the problem of length and the non-uniform structure.


\section{Ethical Considerations}
\label{ethical}
This work conforms to the ethical consideration of the datasets (ILDC \cite{ildc-cjpe} and LexGLUE \cite{lexglue})) used here. 
The framework developed here is in no way to create a "robotic" judge or replace one in real life, but rather to help analyze how LLMs and our hierarchical framework can be applied to long legal documents to predict judgments. 
These methods are not foolproof to predict judgments, and should not be used for the same in real-life settings (courts) or used to guide people unfamiliar with legal proceedings. The results from our framework are not reliable enough to be used by a non-professional to make high-stakes decisions in one's life concerning legal cases.

\section*{Acknowledgements}
\small
This work is supported by the LAWBOT project (ANR-20-CE38-0013) and was performed using HPC resources from GENCI-IDRIS (Grant 2022-AD011013937).
%
%
%
\bibliographystyle{splncs04}
\bibliography{ECIR24}
\end{document}